\documentclass[lettersize,journal]{IEEEtran}
\usepackage{amsmath,amsfonts}
\usepackage{algorithmic}
\usepackage{algorithm}
\usepackage{array}
\usepackage[caption=false,font=normalsize,labelfont=sf,textfont=sf]{subfig}
\usepackage{textcomp}
\usepackage{stfloats}
\usepackage{url}
\usepackage{verbatim}
\usepackage{hyperref}
\usepackage{graphicx}
\usepackage{cite}
\usepackage{xcolor}
\begin{document}

\title{Evacuation Management Framework towards Smart City-wide Intelligent Emergency Interactive Response System}

\author{Anuj Abraham$^{a}$\textsuperscript{\textsection}, Yi Zhang$^{b}$\textsuperscript{\textsection}, and Shitala Prasad$^{c}$\textsuperscript{\textsection} \\
   $^{a}$Technology Innovation Institute, 9639 Masdar City, Abu Dhabi, United Arab Emirates, \\
   $^{b}$Agency for Science, Technology and Research (A*STAR), Institute for Infocomm Research (I2R), Singapore,\\
   $^{c}$Computer Science $\&$ Engineering, School of Mathematics $\&$ Computer Science, IIT Goa, India.
\thanks{Emails: anuj1986aei@gmail.com, yzhang120@e.ntu.edu.sg(Zhang$\textunderscore$Yi@i2r.a-star.edu.sg), shitala@ieee.org. }
}


\maketitle
\def\thefootnote{\textsuperscript{\textsection}}\footnotetext{These authors contributed equally to this work.}\def\thefootnote{\arabic{footnote}}




\maketitle

\begin{abstract}
A smart city solution toward future 6G network deployment allows small and medium-sized enterprises (SMEs), industry, and government entities to connect with the infrastructures and play a crucial role in enhancing emergency preparedness with advanced sensors. The objective of this work is to propose a set of coordinated technological solutions to transform an existing emergency response system into an intelligent interactive system, thereby improving the public services and the quality of life for residents at home, on-road, in hospitals, transport hubs, etc. In this context, we consider a city-wide view from three different application scenes that are closely related to people’s daily life, to optimize the actions taken at relevant departments. Therefore, using artificial intelligence (AI) and machine learning (ML) techniques to enable the next generation connected vehicle experiences, we specifically focus on accidents happening in indoor households, urban roads, and at large public facilities. This smart-interactive response system will benefit from advanced sensor fusion and AI by formulating a real-time dynamic model. 
\end{abstract}


\section{Introduction}
\IEEEPARstart{W}{ith} the rapid expansion and development of smart cities and the growing urban population, the emergency evacuation and preparedness system play a vital role to enhance public services and safety aspects. In the present cutting-edge, the increased use of intelligence, data analytics and innovative cloud technologies based on advanced sensors, internet-of-things (IoT) devices, and vehicle-to-everything (V2X) communication has provided a promising reliable outcome to determine evacuation routes, control, and adjustment of traffic flows. Beyond this, an enhanced interactive system will ensure improvement in residents’ quality of life, towards emerging 6G-V2X and future WiFi systems~\cite{6G}.

In recent years, there have been two main technologies for V2X communications: a) dedicated short-range communication (DSRC)-based vehicular network and b) the cellular-based vehicular network (C-V2X). With the help of V2X technology, prioritized vehicles can communicate with the surroundings and coordinate with traffic signals to reach their destinations as fast as possible. In addition, the combination of above two communication technologies provide a hybrid V2X platform in order to deploy V2X on a large-scale application for automated driving and intelligent mobility~\cite{V2X}.

The use of multi-technology and integrated sensing and communications has enabled heterogeneous vehicular communication, gaining attention and making significant advances in recent research. Advancements in information and communication technology (ICT) have led to efforts to enhance urban facility management by integrating ICT into physical infrastructure \cite{kim2010planning}. Real-time management aims to minimize losses caused by facility failures by relying on accurate spatial and status information \cite{lee2013integrated}. While studies have focused on real-time management of individual facilities, integrated management of diverse city facilities is less explored. Conventional facility management (FM) systems struggle to respond swiftly to emergencies, resulting in damage recovery rather than prevention. In contrast, the proposed intelligent emergency interactive response system detects abnormalities in advance, analyzes geospatial and real-time data, improving facility management and supporting urban services.

Emergency logistics, a vital component of disaster response, encompasses the planning, management, and control of resources to ensure effective support during emergencies. It addresses the challenge of meeting overwhelming demands with limited resources by providing timely assistance to affected populations \cite{caunhye2012optimization}. Optimization-based models are employed to systematically search for optimal evacuation solutions, formulated as mathematical programming problems with objectives like minimizing total evacuation time (TET), congestion time (CT), or maximizing the number of evacuated individuals within a specific timeframe \cite{bayram2016optimization}. Static models, categorized as user equilibrium or system optimal, assume constant network conditions and have limitations in capturing congestion effects. Dynamic models, utilizing dynamic traffic assignment (DTA) methodology \cite{ziliaskopoulos2000linear}, address the dynamic nature of traffic flows during evacuations by considering real-time traffic conditions using techniques such as the cell transmission model (CTM) \cite{daganzo1995cell}. However, applying dynamic models in real-time operations poses computational challenges.

Shelter locations play a crucial role in protecting people during disasters and providing essential resources. Organizations like FEMA \cite{FEMA1988} and the American Red Cross \cite{ARC2002} provide guidance for shelter selection and design, but uncertainties remain until after the disaster. Optimal shelter selection is vital for increasing evacuation rates and minimizing storm damage. Addressing shelter location decisions and traffic assignment together is important for effective evacuation, as handling them separately can lead to sub-optimal outcomes \cite{kongsomsaksakul2005shelter}. Evacuation models need to consider various scenarios and balance conflicting objectives. Incorporating stochastic and robust approaches in these models helps handle uncertainty in planning, considering uncertain parameters and disruptions~\cite{kulshrestha2011robust}.

Overall, the major contributions of the article include to serve for the following goals:

\begin{enumerate}
\item{To optimize the workforce demand required to providing related services based on available data information.}
\item {Integrating supplementary sensors to improve and accelerate emergency response and infrastructure maintenance procedures.}
\item{Send real-time notifications and upload images to relevant departments on any incident/accident detections anywhere.}
\item{Following a household, road, or indoor facility accident, it is crucial to promptly initiate steps for dispatching ambulances, emergency buses, or other lifesaving forces. This intelligent approach helps save lives and ensures smooth traffic flow by timely deploying emergency healthcare services and implementing effective evacuation strategies.}
\end{enumerate}

\section{Scope of Work}
\begin{figure}[!t]
\centering
\includegraphics[width=3.5in]{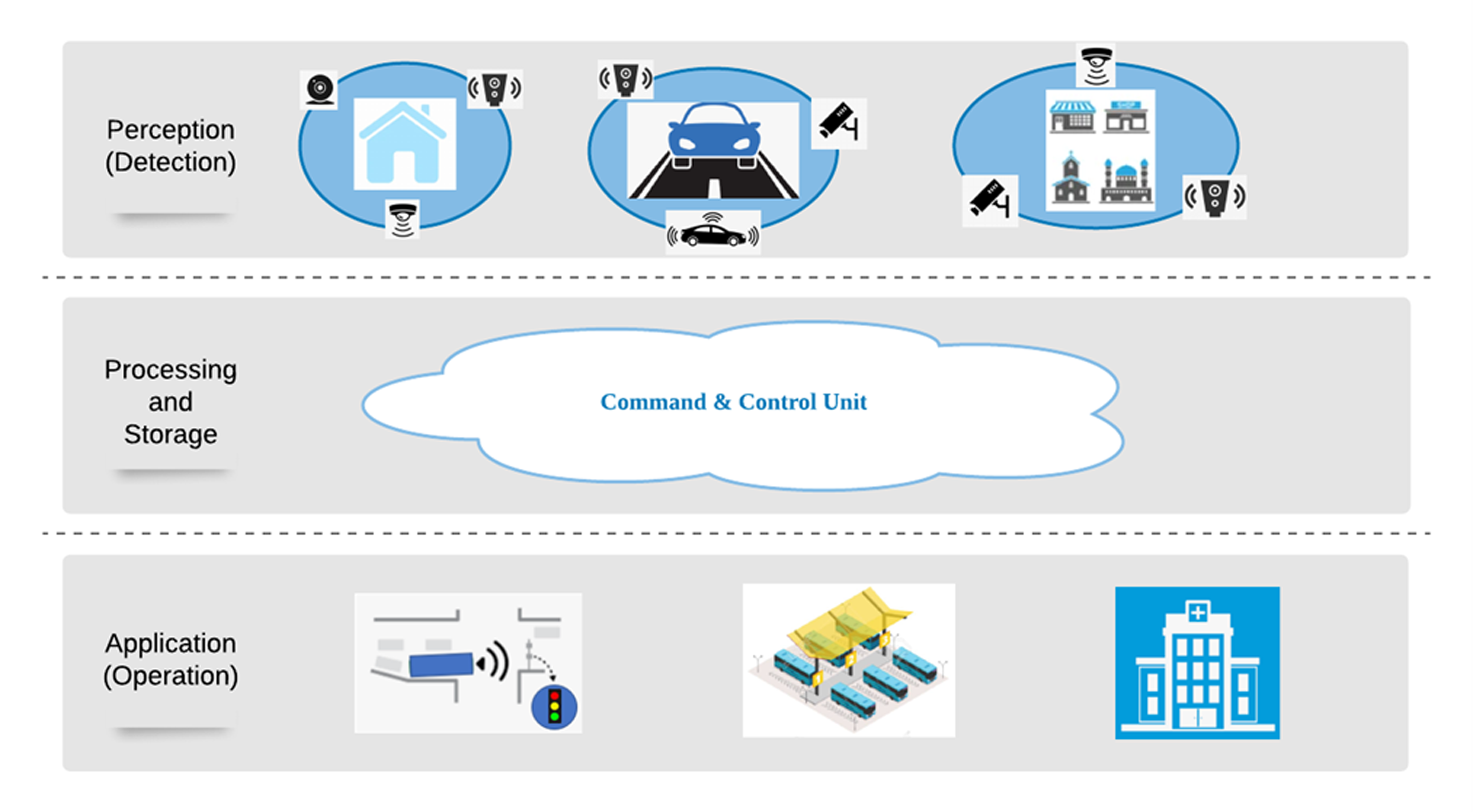}
\caption{Overall stages involved in the smart interactive system.}
\label{fig_1}
\end{figure}

To achieve the above listed goals, we propose a Smart City-wide Intelligent Emergency Interactive Response System to provide a holistic urban fast-response emergency strategy. The proposed idea can be executed in three different stages, as depicted in Figure~\ref{fig_1}. Details can be found as follows:
\subsection{Detection stage for perception}
Initially, an assessment is conducted on the data collection tools currently utilized in the city, utilizing the results to enhance public services and residents' quality of life. This phase primarily emphasizes the careful selection of advanced sensors and the effective gathering of data. Additionally, we compiled a list of potential sensors that can be deployed in indoor households, urban roads, and public facilities.

\subsection{Processing and storage}
The processed data collected from the various sensors are classified into different emergency types and levels. The focus will be on fusing the information from different sensors as well as surrounding infrastructure on-road. This stage also performs a digital platform with real-time notifications that notifies the relevant departments for immediate assistance. Deep learning techniques will be deployed to enable the classification of the emergency levels, and the notification of corresponding departments. Furthermore, the basic information observed from the sensor, such as the location of the accident, the population impacted, the congestion (link flow, average vehicle speed) of around the incident site, will also be incorporated into the notifications and send to related departments. The emergency departments will carry out the next operations based on the received notifications.

\subsection{Application layer for department operations}
As the crucial stage of this proposed system to implement response action, we have developed different optimization models and algorithms to realize a fast-response operation under three different scenes: indoor household accident, urban on-road traffic accident and public facility accident. Optimization schemes for hospital selection and ambulance routing will be introduced to provide immediate medical services for household communities. Also, this proposed system can adapt and extend some existing schemes to help realize various on-road movement enhancement, such as signal priority provided to emergency vehicles, diverting vehicles by re-routing them to uncongested areas. 

\begin{figure}[!t]
\centering
\includegraphics[width=3in]{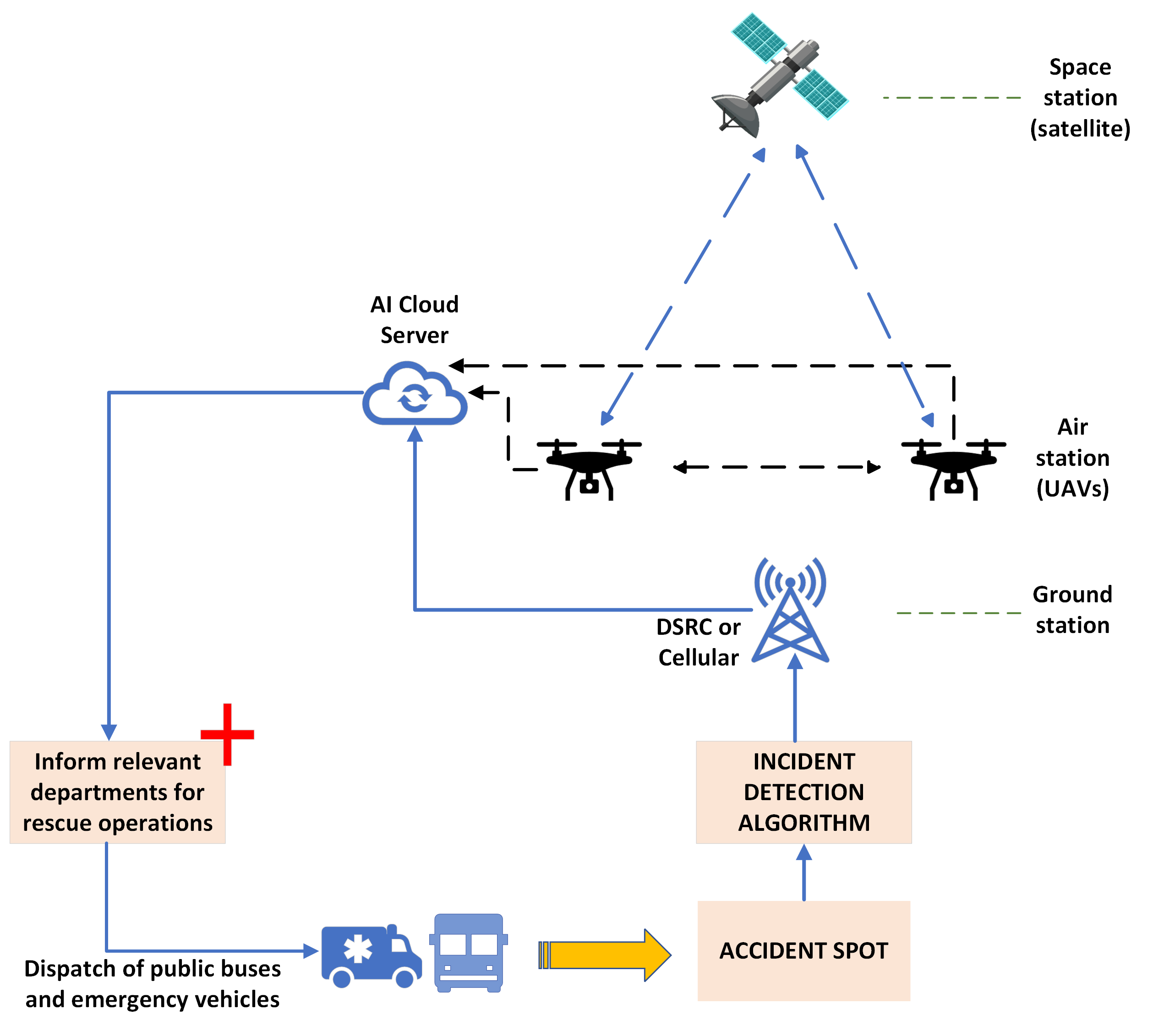}
\caption{The message routing of proposed smart interactive system.}
\label{fig_2}
\end{figure}

A multi-model evacuation system can be developed to improve emergency evacuation around large public facilities. This system includes group-based ambulance dispatching and emergency bus scheduling and routing. The process of message routing, from the detection stage to the operation stage, is illustrated in Figure \ref{fig_2}. Sensors near the accident spot detect abnormal occurrences and send computed messages to a cloud server through communication channels like DSRC or Cellular. Machine learning modules in the server process these messages and notify the relevant departments to take action. Each department then locally determines its operation action using an optimization algorithm.

\section{Industry relevance}
The implementation of smart intelligent systems extends its benefits to industrial safety measures. Utilizing artificial intelligence (AI) in conjunction with surveillance cameras, live footage is continuously analyzed to detect any abnormal activities, incidents, or accidents~\cite{abbr}. Upon detection, the corresponding area is immediately secured, and the necessary assistance is promptly summoned.

In cases where immediate medical attention is required, such as a serious incident involving a person, a coordinated effort with traffic departments establishes a designated pathway through a green door. This facilitates swift access for medical personnel to reach and assist the individual.

Furthermore, the proposed intelligent system proves advantageous in redirecting goods efficiently. By incorporating advanced sensors deployed by small and medium-sized enterprises (SMEs), a comprehensive approach is adopted. These sensors, equipped with intelligent algorithms, continually update and monitor the surroundings in real-time, enhancing security and safety measures.

\section{The Proposed Smart-Interactive Response System}
\subsection{Detection stage}
This section focuses on the proper selection of sensors and efficient data collection from these sensors are to be implemented in corresponding application scene. 

\emph{Major Emergency}: 
To identify individuals experiencing cardiac arrest, unconsciousness, difficulty breathing, seizures, severe injuries, strokes, head trauma, bone fractures, asthma attacks, elderly individuals with chronic health conditions, sick children, cuts with bleeding, accidents resulting in bruising and swelling, minor injuries, and persistent fevers.

\emph{Minor Emergency}: To detect person suffering from constipation, chronic cough, diarrhoea and skin rash.

\subsubsection{Home-based Fall Accident Detection (enhance public health, vital sign triage)}
Enhancing public health and vital sign triage, the implementation of home-based fall accident detection systems addresses the challenges posed by the growing elderly population and the increasing trend of people living alone. This technology also addresses concerns related to leaving infants unattended at home due to busy schedules. By installing advanced in-home detection sensors, the vulnerability of the community can be mitigated, ensuring household preparedness and alerting caregivers through telematics means.

Sensors used to detect emergencies are sound sensors, passive infrared sensor (PIR), virtual sensing for predictive maintenance etc. 


For example, SME sensors provided by SoundEye prototype (in Singapore) which can be used in the detection stage comprise of SoundEye ARK, SoundEye vision, and a smart recessed light LIGHTO etc. (See link: \href{https://sound-eye.com/}{https://sound-eye.com/}) These sensors are planned to deploy for elderly nursing home, airport, research labs, handicap washroom, and many more.


\subsubsection{On-Road Traffic Accident Detection}
A smart traffic management system in urban areas uses V2X communication technology to enhance road safety and reduce traffic congestion. Bluetooth detectors are effective in detecting on-road accidents. Intersections with traffic signals are common bottlenecks in urban commuting. Traditional traffic signals have room for improvement, but integrating V2X communication offers opportunities to develop advanced applications for better traffic signaling.




Unlike traditional segment-based detection methods that rely on expensive technologies such as inductive loops, radar detectors, or local sensors using radar, lidar, laser, infrared, ultrasonic, CCTV cameras, or video technology (scanners), Bluetooth sensors offer a more affordable and accessible alternative. They provide reliable travel time information, making them a preferred choice for monitoring and managing transportation networks efficiently.



Traffic data collection is facilitated by various devices such as cell phones, smartphones, tablets, and Bluetooth-enabled systems within vehicles. These devices have unique MAC addresses that serve as identifiers. To detect Bluetooth devices in urban or freeway settings, dedicated scanners with omnidirectional or directional antennas are installed at fixed locations. These scanners establish a "handshake" with passing vehicles, allowing for information exchange.

Incident detection can be achieved through the use of vibration sensors. In the event of a crash, the sensor activates a circuit that triggers an S.O.S message to nearby hospitals, police stations, fire stations, etc. Additionally, emergency vehicle prioritization can be achieved using the GLOSA algorithm, which adjusts traffic signal timings to facilitate their passage. Furthermore, re-routing based on traffic congestion at intersections can be implemented to optimize emergency response.

\subsubsection{Large Public Facility Abnormal Detection (fire/ terrorist)}


This stage focuses on designing and adapting systems for large public gatherings to interpret information from various surveillance modes, such as CCTV, IR cameras, intrusion detection alarms, and acoustic sensors. It aims to quickly notify infrastructure authorities and relevant departments in case of abnormal detections like fire or terrorist attacks for prompt rescue operations. The proposed strategy includes:
\begin{itemize}
\item Advanced sensors in public facilities and emergency bus stops to monitor evacuee populations.
\item Efficient machine learning algorithms to identify emergency types and levels and notify relevant departments.
\item Group-based ambulance routing to optimize ambulance utilization and minimize patient waiting time.
\item Emergency bus pick-up and delivery strategy for efficient transfer of evacuees to shelters.
\item Signal priority for emergency buses and ambulances to reduce travel time.
\end{itemize}
Additionally, a platform for uploading images is needed to enhance and expedite emergency response and infrastructure maintenance processes for all the discussed scenarios.

\subsection{Data fusion and ML algorithm development stage}


This section describes an intelligent identification system that uses ML algorithms to classify emergency types and levels and notify relevant departments. The system uses data collected from various sensors, such as RGB and near infrared (NIR) cameras. The data is pre-processed and annotated before being used by the ML algorithms.

DL is a leading technology for multimedia analysis, and it can be used to extract features from images and videos that can be used to classify emergencies and their levels. The system also uses an automated notification system to notify relevant departments of emergencies. This notification system is real-time, so that help can be provided as quickly as possible.

\begin{figure}[!t]
\centering
\includegraphics[width=3.5in]{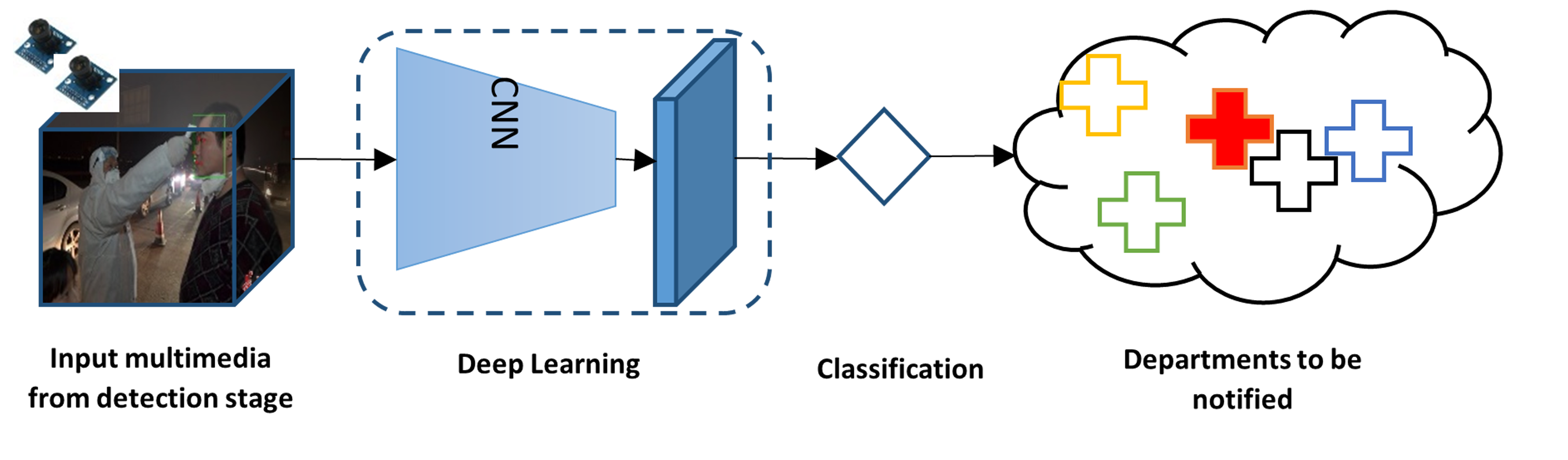}
\caption{AI module to classify incidents/accidents and notifying the relevant departments for assistance.}
\label{fig_3}
\end{figure}


In the depicted system (Figure~\ref{fig_3}), sensor data such as multimedia (images and videos) is captured during the detection stage. This data is then utilized in real-time by the data fusion and ML algorithm development stage. A CNN-based feature extractor is employed to optimize the feature space for representing incidents/accidents. Based on this feature scope, the system determines the type and severity level of the incident/accident and promptly notifies the relevant department for assistance.

The sensor information is pre-processed and essential features are extracted. These stages involve sensors placed in various locations, including homes for the elderly, public roads, and other facilities. Additionally, the system incorporates sensor fusion from vehicle sensors such as cameras, radars, and lidars (used for vehicle localization), as well as external sensors installed along the road network.

The fusion algorithm combines the represented features to classify the emergency type/level. A smart identification system, known as the command and control unit (CCU), then determines the appropriate department to notify based on the received information, such as police station, hospital, or bus terminal/transport hub.

\subsection{Optimized operation stage}
This section focuses on creating and assessing a rapid-response system that integrates optimization models and algorithms in specific application scenarios. The goal is to enhance the efficiency of the current urban response system when it receives operational instructions from the Central Control Unit (CCU) discussed earlier.
\subsubsection{Household emergency service system}
The household emergency service system consists of two integrated optimization problems: the adaptive community covering problem (ACCP) and the dynamic ambulance routing problem (DARP). These problems are solved sequentially to effectively serve the population. In urban cities with multiple hospitals, it is crucial to allocate communities to the appropriate hospitals.

\begin{figure}[!t]
\centering
\includegraphics[width=3in]{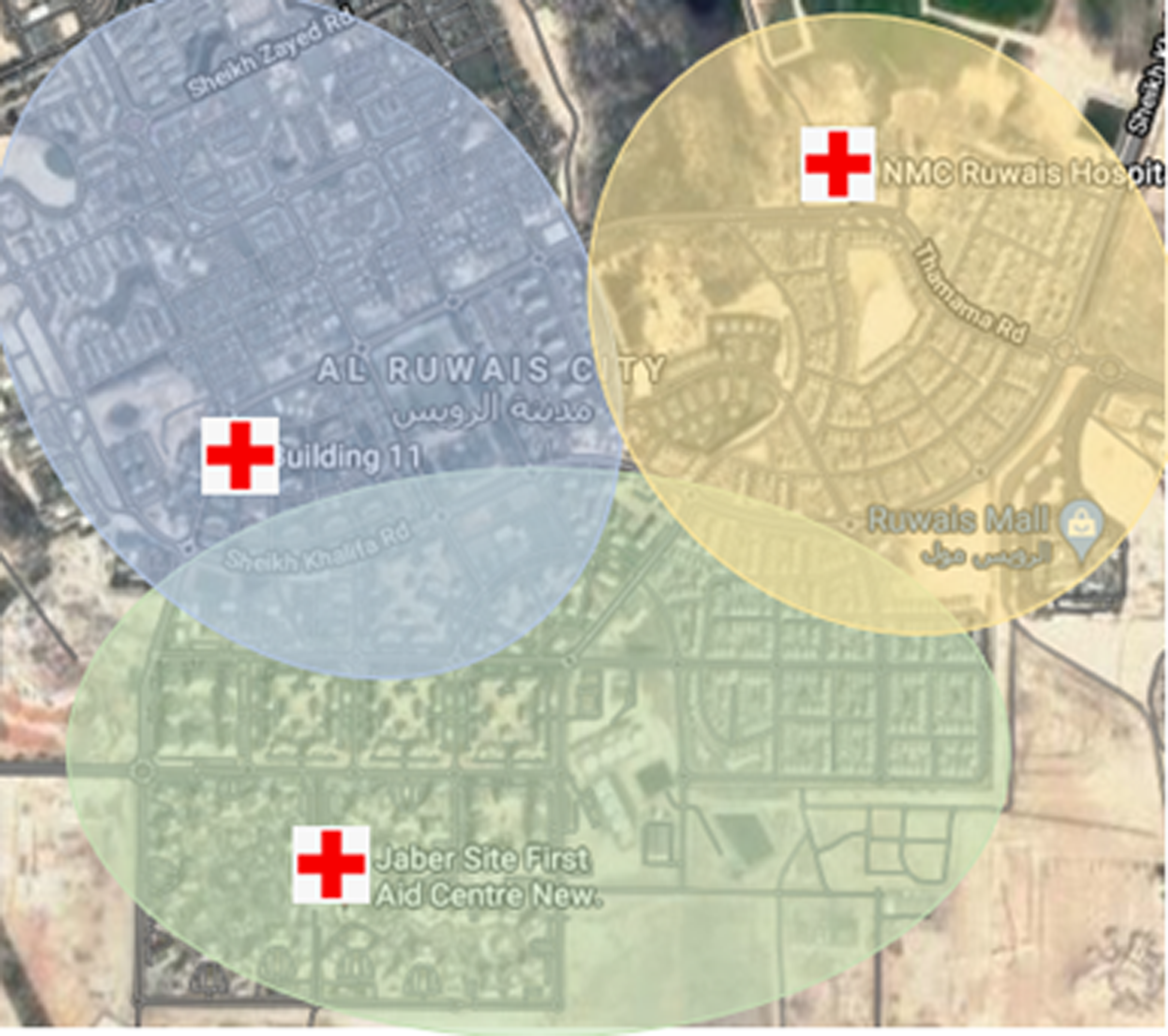}
\caption{Graphical representation of the community covering problem.}
\label{fig_5}
\end{figure}

The community covering problem aims to maximize hospital coverage while minimizing the response time. Figure~\ref{fig_5} illustrates a representation of this problem in Ruwais, where communities within a certain range of the hospital are allocated to it.  It involves assigning communities within a certain range to specific hospitals. By considering real-time factors such as sensor data and current traffic conditions, the ACCP becomes a dynamic model rather than a static one. This allows for the relocation of hospitals based on the current traffic environment, considering time-dependent travel times.

In areas with fluctuating traffic volumes, the proposed method provides better solutions compared to static models. The variation in travel speed and time throughout the day can significantly impact the quality of the solution. Therefore, cities with noticeable traffic variations can greatly benefit from this approach.

\begin{figure}[!t]
\centering
\includegraphics[width=3in]{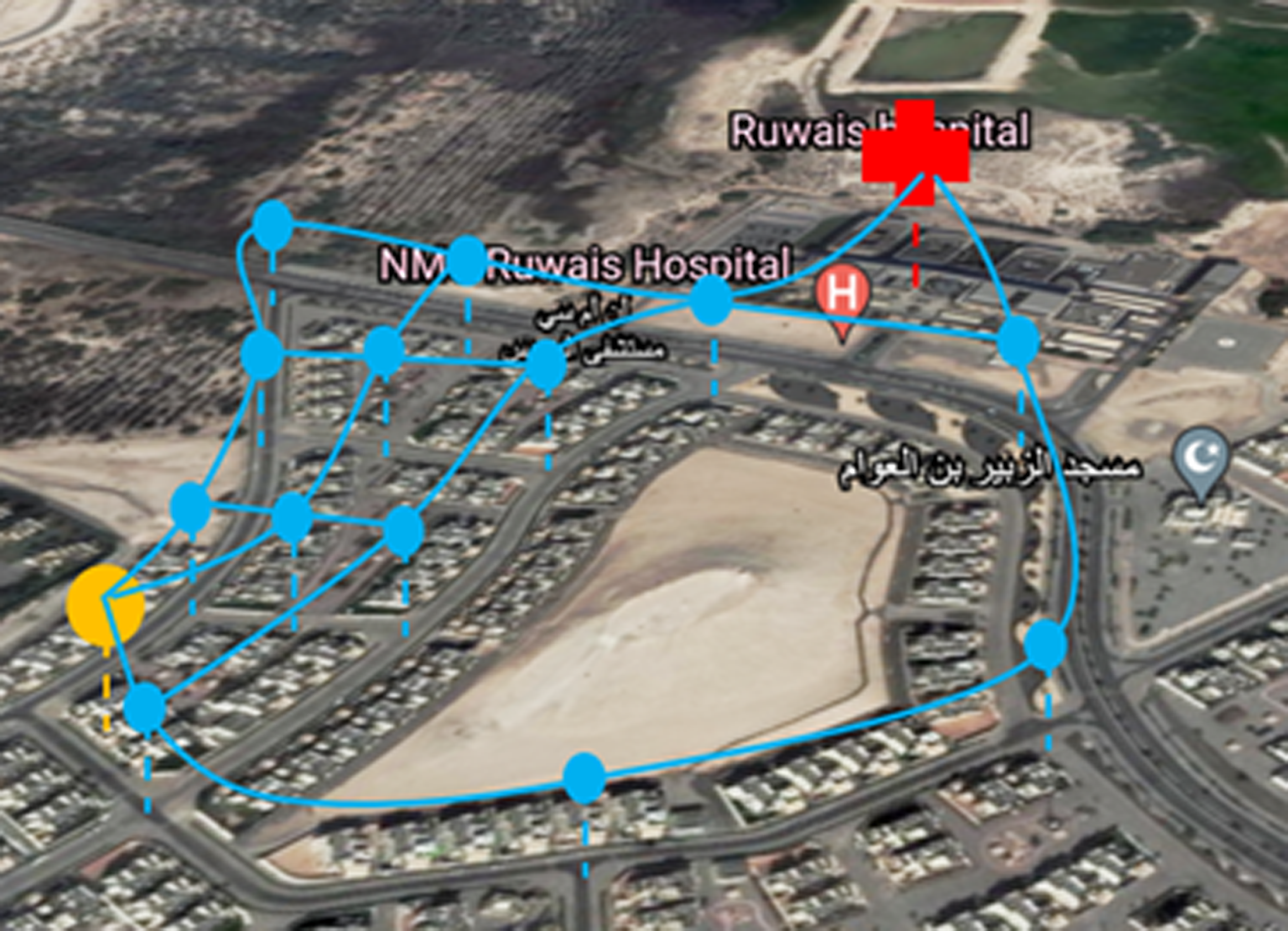}
\caption{Graph representation of network in dynamic ambulance routing.}
\label{fig_6}
\end{figure}

After the determination of the hospital for each community, the routing of the ambulance from the specified hospital for the targeting building in the community comes into the picture. Different with vehicle routing problem (VRP) which serves a group of demands, the household ARP is a single pickup and drop-off problem, and the proposed DARP considers the variations of traffic congestions in computing the shortest path, in other words, the shortest path herein refers to “shortest-time” path. In ACCP, a rough online traveling time from the community to the hospital is used to capture system objectives, while the real-time link-based traveling time is required for the whole network in DARP. Figure~\ref{fig_6} illustrates the road network around NMC Ruwais hospital in UAE, where the red cross, the yellow circle, and blue circles denote the hospital, the demand location, and the intersections, respectively. Although the network structure is fixed, the traveling time on each arc (link) of the network changes throughout the day, which leads to the dynamic update of the shortest path for guiding ambulance. 

The complexity of both ACCP and DARP increases exponentially as the increase of the network scale, the traditional mathematical techniques, such as linear relaxation and branch and bound, cannot efficiently solve the problem in real-time. To address this issue, we will adopt the population-based evolutionary algorithm to smartly traverse the search space via its meta-heuristic. 

\subsubsection{Adaptive signal control and vehicle re-routing to alleviate congestions after traffic accidents}

\begin{itemize}
\item{\emph{GLOSA for emergency vehicle green wave prioritization}:

In this stage, GLOSA applications are tested for different traffic scenarios. Dynamic phase adjustment of the traffic signals can be simulated and combined with the GLOSA algorithm. An emergency vehicle prioritization scenario will be simulated and the effect on the neighboring traffic shall be determined \cite{abraham}. GLOSA uses V2X communication to pass the signal-timing message to the approaching vehicles. 


The GLOSA prediction system uses traffic prediction data along with the model built and provides a green wave prioritization for the emergency vehicle, and the implementation details of vehicle priority order are discussed in the next section (Multi-modal traffic management for a large facility evacuation).}

\item{\emph{Notification display and vehicle re-routing to avoid entering into accident (congestion) area}:
Vehicle prioritization on road networks is becoming popular for completing tasks and smooth maneuvering. Actual traffic data from ground sensors is combined to create relevant traffic information for re-routing vehicles to avoid congested areas. The system architecture includes a learning predictor for data processing at the backend. When an incident is detected, the data from traffic light systems is sent to the predictor, and the congestion level is estimated. The control center receives the incident information and notifies the relevant department. Data analytics provides useful insights. Real-time traffic data from vehicle sensors and traffic light controllers is collected using wireless communication technologies such as DSRC or cellular (4G or 5G or 5G Beyond) and future WiFi system and 6G technologies.

A vehicle sends a route-planning query to the predictor using these wireless communications. The query is handled by the learning predictor application to estimate the shortest travel times. The query contains information such as destination, present location, current speed, and acceleration. The learning predictor uses the query, map, and real-time and historical data collected by traffic light systems to find an optimal route for the vehicles. This route is sent to the vehicle as a suggestion. The route data has information such as path and travel times.

Traffic management based on congestions and incident detections, traffic messages, and estimated travel times are displayed on electronic signboards along the road (freeways/major arterial roads) at specified locations. These traffic messages will assist drivers to re-route their trips to reach their destinations in the shortest time.}
\end{itemize}


\subsubsection{Multi-modal traffic management for a large facility evacuation}

Frequent disasters, either man-made (terrorists attack, traffic accidents, or fire hazard) or natural (flood, hurricane, earthquake), require a well-prepared urban evacuation system with fast-response characteristics to reduce the loss of life as much as possible. Studies have already shown that the severity of a disaster can be largely affected by the efficiency of logistics transportation during the response phase. In this article, although we focus on providing the solution to a man-made accident for the large facility outdoor evacuation, the solution shall still be available to be implemented in other attack situations.

\begin{figure}[!t]
\centering
\includegraphics[width=3in]{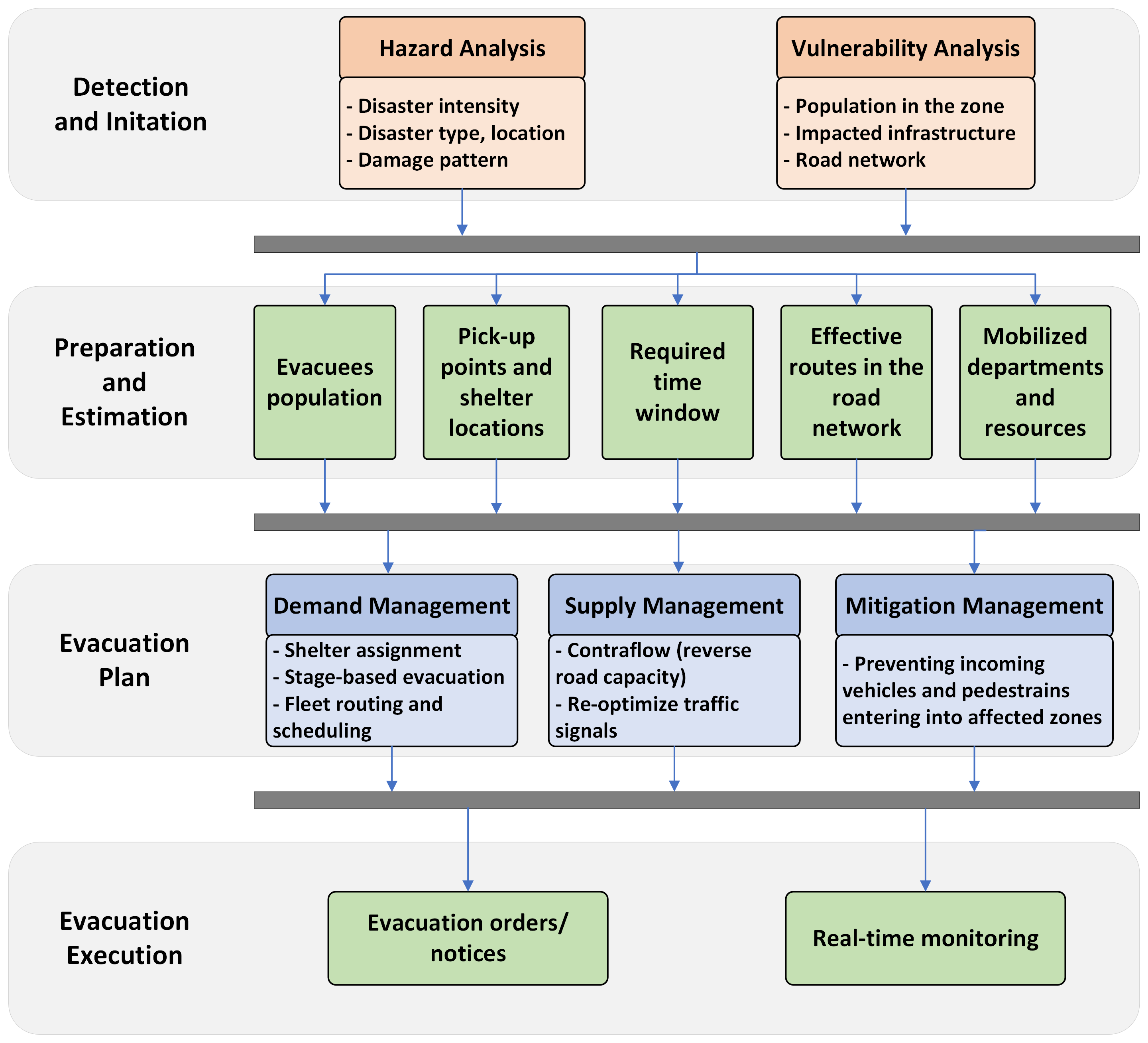}
\caption{The flowchart for the evacuation management framework.}
\label{fig_8}
\end{figure}

The evacuation management system has four major stages:
\begin{itemize}
    \item Sensors collect data about the hazardous area, the number of households affected, and the disaster intensity.
    \item The data is analyzed to estimate the number of evacuees, the pick-up and shelter locations, the evacuation time window, and the accessible routes.
    \item The evacuation plan is generated, which includes demand management, supply management, and mitigation management.
    \item The evacuation plan is implemented by the emergency departments or agencies.
\end{itemize}
Demand management solves the evacuee transfer problems, such as shelter assignment, stage-based evacuee releasement, and vehicle routing to carry evacuees from the hazard zone to shelters. Supply management focuses on providing convenience for transferring evacuees, such as capacity reversibility (contraflow strategy) and adaptive signal control. Mitigation management diverts traffic and sends a warning at the peripheral area to prevent incoming vehicles and pedestrians from entering the hazard zone. Figure~\ref{fig_8} shows the same details.  

In this sub-topic, we propose a multi-modal emergency evacuation strategy. Since the diverting strategy and mitigation action has already been discussed in the previous section of Adaptive signal control and vehicle re-routing, the same approach can be implemented in this evacuation case too. Therefore, we only focus on the demand and supply management herein. Specifically, the demand management considers the ambulance routing for different patient groups and the emergency bus routing to pick up evacuees from the site to shelters, also, the emergency transit signal priority for the supply management. 

\begin{itemize}
\item{\emph{Group-based ambulance routing}:
In a densely populated public facility, dispatching ambulances from a single hospital may not be sufficient to provide first aid for a large number of victims. In this case, ambulances from multiple hospitals or even the entire city may need to be dispatched. To efficiently utilize the ambulances, the patients are divided into two groups: slightly injured and seriously injured. Slightly injured patients can be treated on the field, while seriously injured patients need to be transported to the hospital. The objective of the ambulance routing problem is to minimize the longest waiting time of a patient in both groups. This ensures that the ambulances are used to their maximum capacity and that all patients receive the care they need.}

Figure~\ref{fig_9} illustrates a solution example for this group-based ambulance routing problem, where a shopping mall in Ruwais (located west of Abu Dhabi city in UAE) is selected as an attack object; patients are all located around the attacked building. The red dots and green dots denote the seriously injured patients and slightly injured patients, respectively. The ambulances in three hospitals are dispatched to provide assistance for these patients.
\begin{figure}[!t]
\centering
\includegraphics[width=3in]{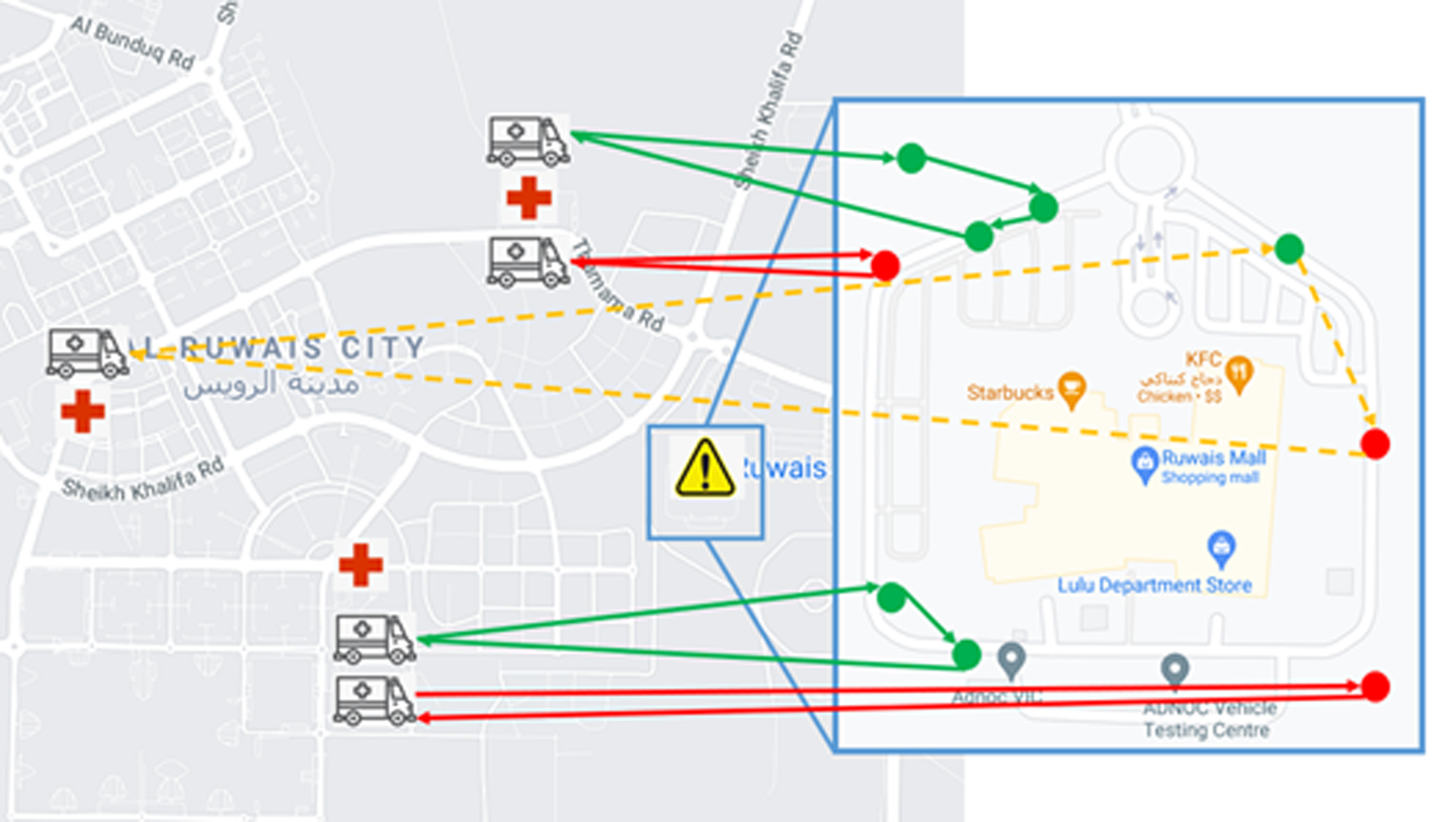}
\caption{Graph example for the ambulance routing solution.}
\label{fig_9}
\end{figure}
\item{\emph{Optimal routing for emergency buses}:
After an emergency happens, the pick-up points within the dangerous zone will be incorporated into the monitoring scope, and the corresponding emergency bus scheduling and routing will be generated according to the evacuee demands around the incident cite. The entire evacuation network is composed by the bus terminals, pick-up points, and shelters, which leads to a multiple pick-up and delivery problem.

Public buses are a good choice for transferring evacuees during an emergency. The city has emergency bus terminals, specified pick-up points, and shelters scattered throughout the city. The pick-up points are equipped with sensors to detect the number of evacuees in real time. After an emergency happens, the pick-up points within the dangerous zone are incorporated into the monitoring scope, and the corresponding emergency bus scheduling and routing are generated according to the evacuee demands. The entire evacuation network is composed of bus terminals, pick-up points, and shelters, which leads to a multiple pick-up and delivery problem.

The Figure~\ref{fig_10} shows an example of an EBPD problem. The triangle, circle, and square denote the bus depots, pick-up points, and shelters, respectively. The blue ellipse indicates the affected dangerous zone, and the size of the circle represents the demand quantity at corresponding bus stops. The number of buses dispatched from each depot, the routing of each bus specifying the visited pick-up points, the evacuation time window constraint, and the boarding control applied at each pick-up point are all factors that need to be considered in the EBPD problem.}

\begin{figure}[!t]
\centering
\includegraphics[width=3in]{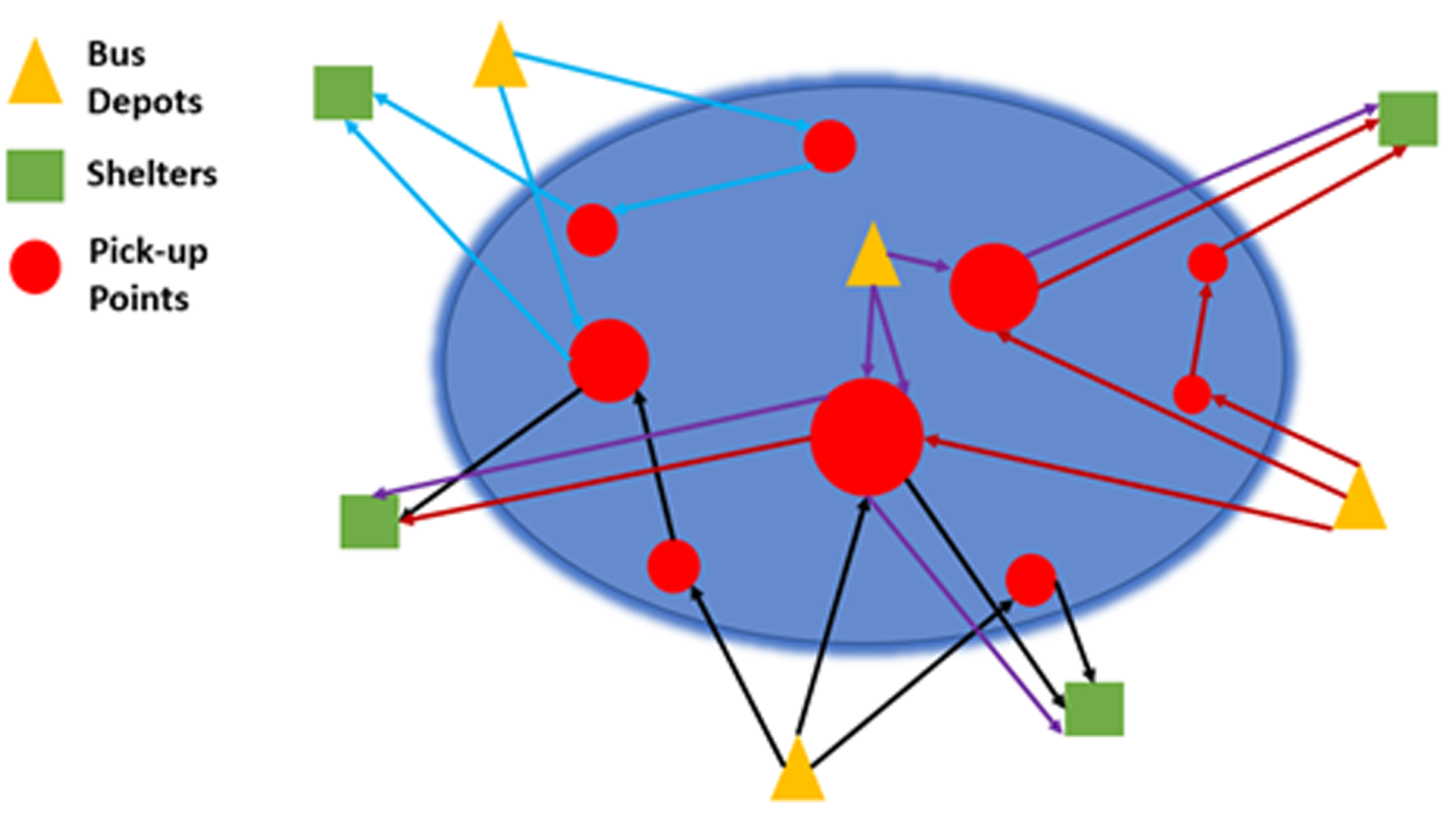}
\caption{Graph example for the emergency bus pick-up and delivery solution.}
\label{fig_10}
\end{figure}

\item{\emph{Emergency transit signal priority}:


Although transit signal priority has been discussed and deployed in earlier studies \cite{eom2020traffic}, our strategy is targeting on providing convenience to vehicles under emergency situation, the switch of the signal phase shall be more determined than other situations. Also, the priority order assigning to the conflict emergency vehicles approaching the intersection from different approaches shall also be included in the study. Attributes, such as bus demand, shelter distance, and number of remaining pick-up points required to be visited by this bus, will all be considered into the assessment system to evaluate the priority order.

Supply management is a way to smooth the flow of emergency vehicles on the road. Emergency vehicles with onboard units (OBUs) installed can communicate with road-side units (RSUs) in real time. The online information gathered at RSU will be transmitted to signal controller to execute corresponding control schemes. 
The emergency vehicle is provided with the highest priority. No matter which phase is operating, the current phase will change to the targeting phase to provide right-of-way for the emergency vehicle. If the current phase is just the targeting phase, its duration will be extended till the emergency vehicle safely crosses the intersection.}
\end{itemize}





\section{Conclusion and Research Directions}

We propose a smart-interactive response system for emergency preparedness that focuses on accidents in indoor households, urban roads, and large public facilities. Our solution uses advanced sensors and V2X technology to connect SMEs, industry, and government entities with infrastructure.
In the event of an accident or incident, we propose a structured approach to dispatching ambulances, emergency buses, or other lifesaving forces. This approach can be intelligently initiated to save lives and clear roads for traffic flow. AI/ML, including deep learning, is highly accurate and advanced for assisting in such incidents.
For intelligent transportation mobility systems, we discuss the potential of AI/ML-enabled V2X communication to improve public services and safety for residents. This includes re-routing methods, traffic signaling control, and other technological solutions for emergency evacuation.

In future work, we will explore the use of federated learning to ensure user privacy in transmitting wireless signals, and to enable collective intelligence via 6G V2X and 6G IoT networks for a safe evacuation. Also, to integrate global 3-tier communication network coverage (unmanned aerial vehicles (UAVs), earth stations, and satellites) will be more focused on a city-wide solution toward disaster management framework.



%
\bibliographystyle{IEEEtran}
\bibliography{ref.bib}




\newpage

\section{Biography Section}
\vspace{1pt}
\begin{IEEEbiography}[{\includegraphics[width=1in,height=1.25in,clip,keepaspectratio]{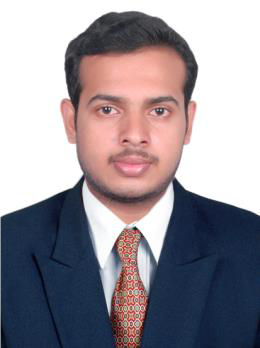}}]{Anuj Abraham} is currently a Senior Researcher in AI and Digital Science Research Center (AIDRC) at Technology Innovation Institute, Abu Dhabi, United Arab Emirate. Previously, he was a Scientist at the Institute for Infocomm Research (I$^{2}$R), a research entity of the Agency for Science, Technology and Research (A*STAR) in Singapore.  He has several years of expertise in the field of control engineering and smart mobility.
\end{IEEEbiography}

\begin{IEEEbiography}[{\includegraphics[width=1in,height=1.25in,clip,keepaspectratio]{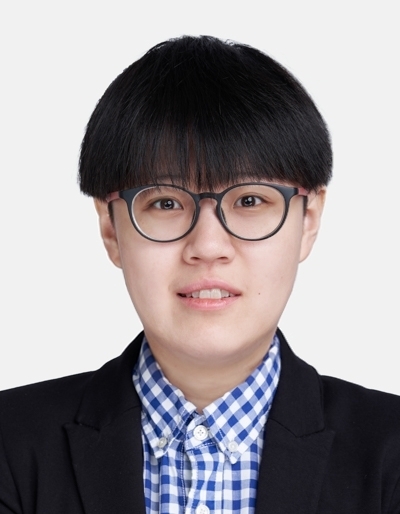}}]{Yi Zhang} is currently a Scientist at the Institute for Infocomm Research (I$^{2}$R), a research entity of the Agency for Science, Technology and Research (A*STAR) in Singapore. She received her PhD degree in Electrical and Electronic Engineering from Nanyang Technological University, Singapore 2020. Her research interests focus on operation research and control on the application of intelligent transportation system.
\end{IEEEbiography}

\begin{IEEEbiography}[{\includegraphics[width=1in,height=1.25in,clip,keepaspectratio]{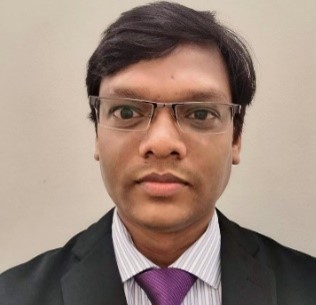}}]{Shitala Prasad}
is an Assistant Professor at IIT Goa with a PhD from IIT Roorkee in computer vision area. He is an expert in image analysis and understanding, with applications in biometrics, agriculture, biomedical and industry. He has worked at I2R Singapore, NTU Singapore, and GREYC-CNRS France. He is a Senior IEEE Member and has been recognized for his contributions to the scientific community.
\end{IEEEbiography}

\vspace{11pt}

\vfill

\end{document}